\documentclass[conference]{IEEEtran}
\IEEEoverridecommandlockouts
\usepackage{cite}
\usepackage{amsmath,amssymb,amsfonts}
\usepackage{algorithmic}
\usepackage{graphicx}
\usepackage{textcomp}
\usepackage{xcolor}
\def\BibTeX{{\rm B\kern-.05em{\sc i\kern-.025em b}\kern-.08em
    T\kern-.1667em\lower.7ex\hbox{E}\kern-.125emX}}
\begin{document}

\title{SSM-Net for Plants Disease Identification in Low Data Regime
}

% \author{\IEEEauthorblockN{Removed for blind review}}
\author{\IEEEauthorblockN{Shruti Jadon}
\IEEEauthorblockA{\textit{IEEE Member} \\
% \textit{University of Massachusetts, Amherst}\\
shrutijadon@ieee.org \\
https://orcid.org/0000-0002-6953-142X}
}
\maketitle

\begin{abstract}
Plant disease detection is an essential factor in increasing agricultural production. Due to the difficulty of disease detection, farmers spray various pesticides on their crops to protect them, causing great harm to crop growth and food standards. Deep learning can offer critical aid in detecting such diseases. However, it is highly inconvenient to collect a large volume of data on all forms of the diseases afflicting a specific plant species. In this paper, we propose a new metrics-based few-shot learning SSM net architecture, which consists of stacked siamese and matching network components to address the problem of disease detection in low data regimes. We demonstrated our experiments on two datasets: mini-leaves diseases and sugarcane diseases dataset. We have showcased that the SSM-Net approach can achieve better decision boundaries with an accuracy of 92.7\% on the mini-leaves dataset and  94.3\% on the sugarcane dataset. The accuracy increased by ~10\% and ~5\% respectively, compared to the widely used VGG16 transfer learning approach. Furthermore, we attained F1 score of 0.90 using SSM Net on the sugarcane dataset and 0.91 on the mini-leaves dataset. Our code implementation is available on Github: https://github.com/shruti-jadon/PlantsDiseaseDetection.
\end{abstract}


\section{Introduction}

With an exponential population growth rate, it has become critical to produce sufficient food to meet global needs. However, food production is a complex process involving many issues such as climate change, soil pollution, plant diseases, etc. Plant diseases are not only a significant threat to food safety on a global scale but also a potential disaster for the health and well-being of consumers \cite{singh2018pesticide} \cite{hoppin2017pesticides}. Due to the difficulties involved in proper disease identification, farmers apply a mixture of various pesticides, which in turn causes vegetation loss, subsequently leading to either monetary loss or affecting health. In the USA alone, food allergy cases have increased by approximately 18 percent since 2003. \cite{hoppin2017pesticides} By leveraging the increase in computing power we can take advantage of deep learning methodologies for disease detection. Nevertheless, in many scenarios, it is almost impossible to collect a large volume of data concerning a particular disease in a plant species. For example, sugarcane fungal infections such as red rot (Colletotrichum falcatum) and smut (Sporisorium scitamineum) are predominant in the South Indian peninsula. This can lead to a highly unbalanced dataset.

Considerable work \cite{badage2018crop}\cite{Saleem2019}\cite{Nagasubramanian2019} \cite{Bhatia2020} has been done to address the problem of plant disease identification \cite{maniyath2018plant} \cite{Wang2017} using machine learning, but no approach has thus far been proposed to tackle it in low data regimes.

\begin{figure}[!ht]
\begin{center}
   \includegraphics[width=1.0\linewidth]{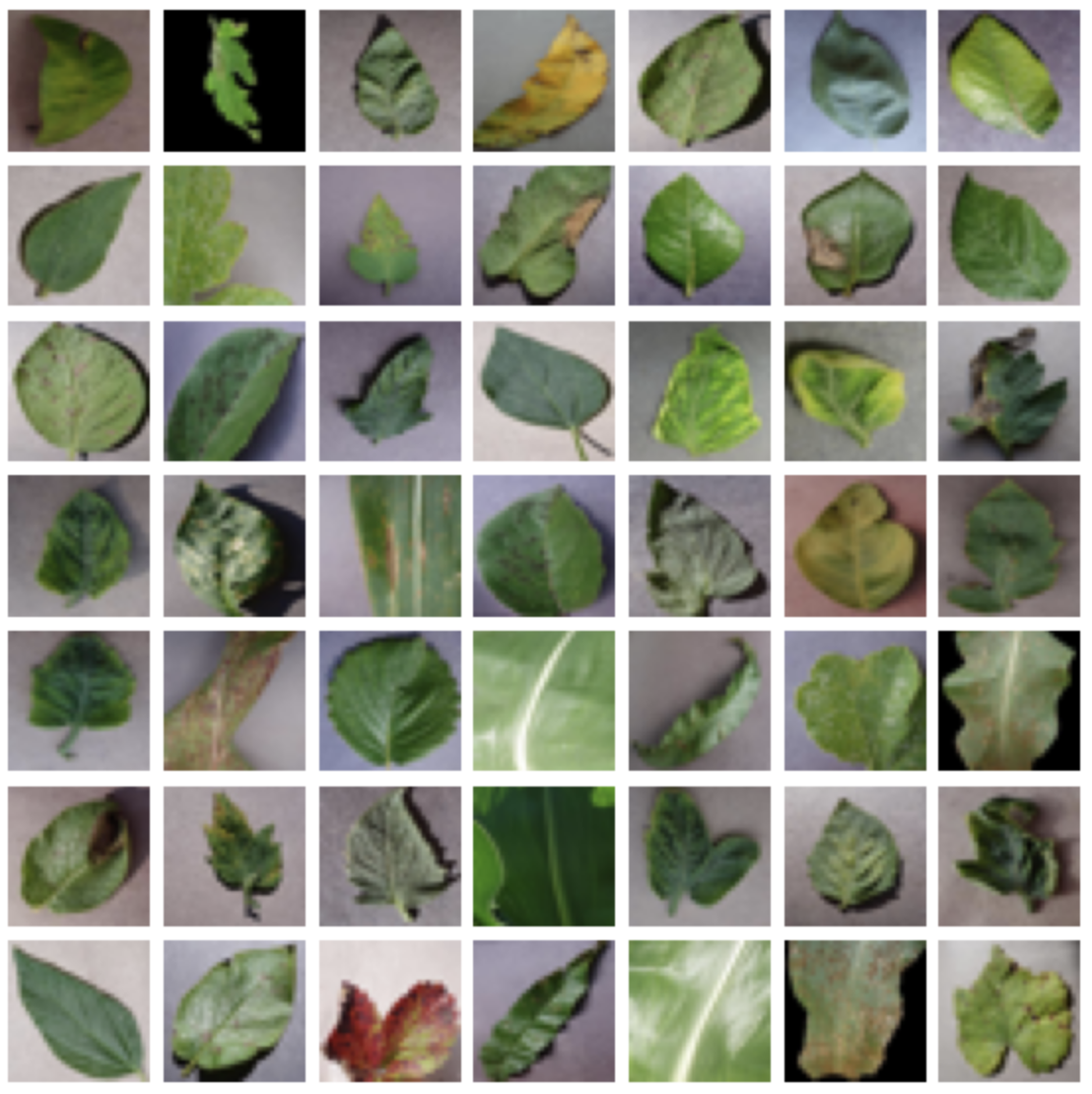}
\end{center}
   \caption{Sample Images of types of Leaves Diseases in Mini-Leaves dataset \cite{aicrowd} by AI Crowd}
\label{fig:0}
\end{figure}

However,  Mrunalini et al.\cite{badnakhe2011application} used traditional vision based feature extraction techniques followed by clustering using K-means to accurately classify leaf diseases. Similarly, Piyush et al. \cite{chaudhary2012color} has proposed an ensemble image filter followed by spot segmentation, which doesn't require huge amount of data for training. The only drawback of using traditional vision based filters for feature extraction is that there's no universal approach, unlike Convolutional neural network architectures, which works for various objectives. Few-shot learning concept has recently grabbed attention of a lot of AI applications such as imitation learning, medical disease detection etc. Few-shot as concept have various implementation ranging from non-parametric approaches such as K-Nearest Neighbor to complex deep learning algorithms. 

In this paper, we propose SSM-Net, a few-shot metrics-based deep learning architecture for plants disease detection in an imbalanced and scarce data scenario. We have compared our proposed SSM-Net with other well known metric-based few-shot approaches in terms of accuracy and F1-score metric. Furthermore, we have showcased that by using proposed SSM Net (Stacked Siamese Matching), we have been able to learn better feature embeddings, achieve an accuracy of 94.3\% and F1 score of 0.90. The code is available on Github:https://github.com/shruti-jadon/PlantsDiseaseDetection.

\begin{figure}[t]
\begin{center}
% \fbox{\rule{0pt}{2in} \rule{0.4\linewidth}{0pt}
  \includegraphics[width=1.0\linewidth]{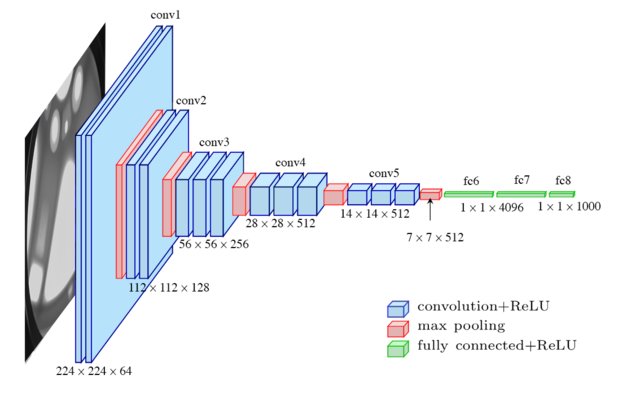}
\end{center}
  \caption{Architecture of VGG16 \cite{simonyan2014very} Network. We took advantage of Transfer Learning(VGG16) to extract better differentiating features up till convolutional layer.}
\label{fig:2}
\end{figure}

\section {Approaches}
Few-shot learning is a method to train an efficient machine learning model to predict any objective with less amount of data. Few-Shot Learning approaches \cite{jadon2019hands} can be widely categorized into 3 cases: Metrics based methods, Models based methods, and Optimization-based methods. In this paper, we have decided to tackle the problem of plant disease detection using metrics-based approaches and compared it with the widely-used transfer learning approach in scarce data cases. Metrics-based methods, as the name suggests, are based upon metrics such as feature embeddings, objective function, evaluation metric, etc. A metric plays a very important role in any Machine Learning model, If we are able to somehow extract proper features in initial layers of a neural network, we can optimize any network using only a few-examples. In this paper, we have taken advantage of two such metrics-based approaches: Siamese Network and Matching Network to create SSM Net. We have also taken the widely used Transfer Learning approach into account and showcased the comparison among all methods in Experiments Section. Before proceeding to Experiments, Let's first understand existing and proposed approaches.

\subsection{Transfer Learning}
Transfer learning \cite{weiss2016survey} refers to the technique of using knowledge gleaned from solving one problem to solve a different problem. Generally, we use the help of well-known networks such as Alex Net, VGG 16, Inception, Exception, etc., trained on the ImageNet dataset. For our case, we have extracted middle layer features of the VGG16 network(refer fig \ref{fig:2}) and fine-tuned by adding a linear layer using a cross-entropy loss function. We have also taken into account that Transfer Learning extracted features can be helpful in learning more advanced objectives and therefore used them to improve upon other approaches as shown in Experiments and Results Section.

\subsection{Siamese Networks}
A Siamese network \cite{koch2015siamese}, as the name suggests, is an architecture with two parallel layers. In this architecture, instead of a model learning to classify its inputs using classification loss functions, the model learns to differentiate between two given inputs. It compares two inputs based on a similarity metric and checks whether they are the same or not. This network consists of two identical neural networks, which share similar parameters, each head taking one input data point. In the middle layer, we extract similar kinds of features, as weights and biases are the same. The last layers of these networks are fed to a contrastive loss function layer, which calculates the similarity between the two inputs.
\begin{figure}[t]
\begin{center}
% \fbox{\rule{0pt}{2in} \rule{0.4\linewidth}{0pt}
   \includegraphics[width=1.0\linewidth]{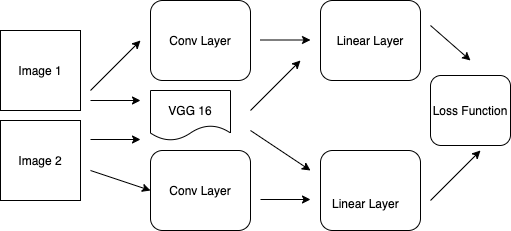}
\end{center}
   \caption{Architecture of Siamese Network. We took advantage of Transfer Learning(VGG16) to extract better differentiating features using Contrastive Loss Function.}
\label{fig:3}
\end{figure}

The whole idea of using Siamese architecture \cite{Jadon2019ImprovingSN}\cite{jadon2019hands} is not to classify between classes but to learn to discriminate between inputs. So, it needed a differentiating form of loss function known as the contrastive loss function. For our case, we have leveraged transfer learning as shown in Fig \ref{fig:3}, to extract complex embeddings which were not possible to learn with less amount of data-set.

\begin{figure}[t]
\begin{center}
% \fbox{\rule{0pt}{2in} \rule{0.4\linewidth}{0pt}
   \includegraphics[width=1.0\linewidth]{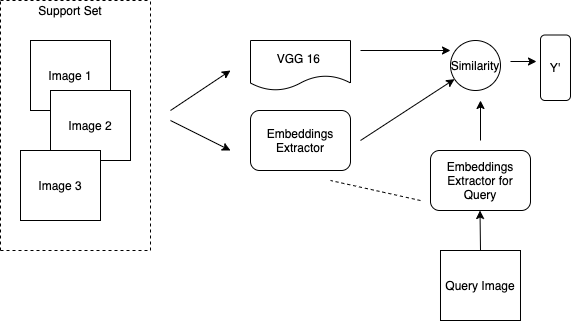}
\end{center}
   \caption{Architecture of Matching Networks. We took advantage of Transfer Learning(VGG16) in process of creating full-contextual embeddings}
\label{fig:4}
\end{figure}

\subsection{Matching Networks}
 Matching networks \cite{vinyals2016matching}, in general, propose a framework that learns a network that maps a small training dataset and tests an unlabeled example to the same embeddings space. Matching networks aim to learn the proper embeddings representation of a small training dataset and use a differentiable kNN with a cosine similarity measure to ensure whether a test data point is something ever to have been seen or not. Matching networks(refer fig \ref{fig:4}) are designed to be two-fold: Modeling Level and Training Level. At the training level, they maintain the same technique of training and testing. In simpler terms, they train using sample-set, switching the task from minibatch to minibatch, similar to how it will be tested when presented with a few examples of a new task. 
 
 At the modeling level, Matching networks takes the help of full-contextual embeddings in order to extract domain-specific features of the support set and query image. For our case, to extract better features from the support set and query image, we have leveraged transfer learning by pre-training on Matching networks on miniImageNet \cite{russakovsky2015imagenet}.

  \subsection{SSM(Stacked-Siamese-Matching) Net}
\begin{figure}[t]
\begin{center}
% \fbox{\rule{0pt}{2in} \rule{0.4\linewidth}{0pt}
   \includegraphics[width=1.0\linewidth]{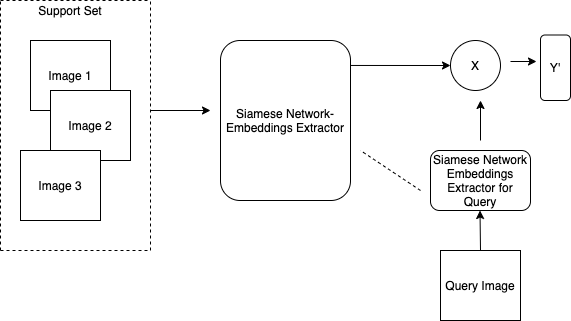}
\end{center}
   \caption{Framework of SSM(Stacked-Siamese-Matching) Net. Here, fine-tuned Siamese Network is being used as discriminative feature extractor plugin on top of Matching Network Architecture.}
\label{fig:5}
\end{figure}

Even for Matching Networks to train and learn better features, we need a decent amount of data to avoid overfitting. Using Siamese Networks we were able to extract good discriminative features. We then decided to leverage these extracted features to learn further about differences among diseases. Therefore, we have proposed a Siamese Head Plugin on top Matching Networks(refer fig \ref{fig:5}) to extract more focused features as shown in Figure 4. In this Network Architecture, instead of extracting features directly from the Transfer learning head, first, we fine-tune them using Siamese Network Architecture(With Transfer Learning Extracted Features) and once the Siamese Network is trained. We extract features using Siamese Networks for our Sugarcane Disease data and feed into pre-trained Matching Networks Architecture on miniImageNet \cite{russakovsky2015imagenet}. Using this approach, we were able to further improve the classification accuracy by ~2\%.
 
\section{Experiments and Results}
In this section, we describe our experimental setup. An implementation of Siamese Network, Matching Network, and SSM-Net is publicly available at Github: https://github.com/shruti-jadon/PlantsDiseaseDetection. 

\subsection{Datasets}
For this work, we have experimented on two datasets: Mini-Leaves data-set \cite{aicrowd} by AI Crowd and our collected sugarcane data-set. Mini-Leaves dataset consists of 43525 training and 10799 test images of plant leaves at 32x32 pixels. These images belongs to 38 types of leaves disease classes. One of the major challenge is the leaves images are distorted, making it complex to extract features. On the other hand, Sugracane dataset is collected with the help of farmers in India. Overall, the aim of this work is to provide farmers (using drones) ability to mass detect disease and spray accurate pesticides in that region. Our data consist of a total of ~700 images of 11 types of sugarcane disease, as shown in Table \ref{table1}.

\begin{table}[!ht]
\begin{center}
\begin{tabular}{|l|c|}
\hline
Disease Type & No. of Images \\
\hline\hline
grassy Shoot & 90\\
leaf spot & 80\\
leaf scald & 46 \\
red rot &  93 \\
nitrogen abundance & 84\\
orange rust & 57\\
pyrilla & 66 \\
smut & 38\\
woolly aphid & 59\\
wilt & 45\\
yellow leaf disease & 42\\

\hline
\end{tabular}
\end{center}
\caption{Categories \& Number of Images of Sugarcane Diseases Dataset}
\label{table1}
\end{table}

\begin{table*}
\begin{center}
\begin{tabular}{|l|c|c|c|}
\hline
Dataset & Method & Augmented & Silhouette-Score \\
\hline\hline
Mini-Leaves Dataset & Transfer Learning & N & 0.064 \\
& Siamese Networks & N & \textbf{0.528} \\
& Siamese Networks[Modified]&N & 0.41 \\
\hline
Sugarcane Dataset & Transfer Learning &N &  0.108 \\
& Transfer Learning &Y & 0.103 \\
& Siamese Networks &N & 0.335 \\
& Siamese Networks &Y & 0.428\\
& Siamese Networks[Modified]&N &  0.341\\
& Siamese Networks[Modified] &Y & \textbf{0.556} \\
\hline
\end{tabular}
\end{center}
\caption{Decision Boundary(Discriminative Features) evaluation using Silhouette Score on Siamese Network and Transfer Learning approach.}
\end{table*}

\subsection{Implementation Details}
 As part of our experiments, we needed to implement four Networks: Transfer Learning (VGG16), Siamese Network, Matching Network, and SSM-Net(proposed). For Transfer Learning, we have used VGG16 Net \cite{simonyan2014very} pretrained on Image-Net dataset. Though VGG16 Net consists of 16 layers, but for our experiments we have extracted features till convolutional layers to avoid learning Image-Net specific features. We have implemented four conv-layered  \textbf{Siamese Network} following \cite{Jadon2019ImprovingSN} with contrastive loss function, It has been modified to take features from VGG16 based transfer learning. We also implemented \textbf{Matching Network} following \cite{jadon2019hands} with LSTM-based embeddings extraction. To ensure the stability of implemented architectures, we have trained Matching Networks and SSM-Net on miniImageNet dataset \cite{russakovsky2015imagenet} and fine-tuned on above mentioned datasets. We use the standard split on miniImageNet dataset of 64 base, 16 validation and 20 test classes. 
 For mini-leaves dataset, we had ~43525 images, we split the data into 25 base classes and 13 test classes, with images of size 32 $\times$ 32 $\times$ 3. Similarly,
 as part of Sugarcane dataset, we had ~700 images, and with data augmentation \cite{wong2016understanding} \cite{cap2020leafgan}, we increased it to ~850 images. The sugarcane dataset is split into 6 base classes and 5 test classes, with images of size 224 $\times$ 224 $\times$ 3.

\subsection{Evaluation Protocol}
We compared our proposed SSM-Net outcomes in terms of decision boundaries, accuracy, and F1-Score. To assess decision boundaries, we have used Silhouette score, widely used in unsupervised learning approaches to evaluate clustering. Silhouette score measures the similarity of an point to its own cluster (intra-cluster) compared to other clusters (inter-cluster). It ranges from $-1$ to $+1$, where a high value indicates better clusters.
\begin{equation}
silhouette-score=\frac{(b - a)}{max(a, b)}
\end{equation}
Here, $a$ is mean intra-cluster distance and $b$ is the mean nearest-cluster distance for each sample.

Similarly, to assess classification outcomes, we have used F1-Score apart from accuracy to assess the quality of our experiments in terms of true positives and false negatives.
\begin{equation}
    f1-score = \frac{2*(precision*recall)}{(precision+recall)}
\end{equation}

For training purposes, we have trained our Matching Networks and SSM-Net as 5 Shot-5 Way i.e; each batch consists of 5 classes per set, and each class has 5 examples.
\subsection{Results}
We compared the original Siamese network and Matching network results with baselines(Transfer learning using VGG16), to validate the effectiveness of metrics-based learning approaches method for classification/identification in the extremely low-data regime. 
\textbf{Note:} We have used Siamese Network for embeddings extraction, not for classification. For classification, we have compared Matching Networks and our proposed SSM-Net.

\textbf{Comparison of Decision Boundaries with Strong Baseline}
We showcased the results of Siamese Networks in comparison to the fine-tuning of VGG16 Network in terms of decision boundaries in Table 2. Note that for decision boundary evaluation, we have extracted last layer embeddings and cluster them to the number of classes i.e; eleven. To evaluate our cluster strength, we have used the silhouette score which calculate inter-cluster vs intra-cluster distance. It is known that the silhouette score of close to 1 means better-defined clusters. For data augmentation techniques we used brightness, random scaling, rotation, and mirror flipping. It is observed that Siamese Network performs well on defining better decision boundaries in comparison to Transfer Learning+Fine-Tuning approach even with data-augmentation. Our modified Siamese Network is able to achieve 0.55 Silhouette score an increase of ~0.45 from Transfer learning with Augmentation on sugarcane dataset. Similarly using mini-leaves dataset, we observed Siamese Network based features resulted in 0.52 Silhouette score a drastic improvement from 0.06 obtained using Transfer learning. We have also noticed that mini-leaves dataset perform better without help of transfer learning features, e.g; in case of Siamese Networks with VGG 16, we obtained 0.41 Silhouette score whereas simple Siamese Networks obtained best outcome of 0.52. 

\textbf{Comparison of Accuracy and F1-Score with Strong Baseline}
Here, we showcased the outcomes of 5 Shot-5 Way SSM Net, Matching Networks, and Transfer learning(VGG16) in terms of Accuracy and F1-Score listed in Table 3. Similar to our last experiment, we have used Data Augmentation techniques. We have observed that our proposed Matching Networks with Siamese Network Head performs better than other approaches. It is able to achieve an accuracy of 94.3\% and an F1-Score of 0.90 on sugarcane dataset. Similarly, on mini-leaves dataset we obtained accuracy of 91\% and F1 score of 0.85 using SSM-Net. 

\begin{table*}
\begin{center}
\begin{tabular}{|l|c|c|c|c|}
\hline
Datset & Method & Augmentation & Accuracy & F1-Score \\
\hline\hline
Mini-Leaves Dataset & Transfer Learning[VGG]& N &  77.5\% & 0.693 \\
& \textbf{Matching Networks}& N & \textbf{84.5\%} &0.83 \\
& Matching Networks[miniImageNet]& N &82.7\%&0.80 \\
& \textbf{SSM-Net}& N&\textbf{92.7\%}&\textbf{0.91} \\
\hline
Sugarcane Dataset & Transfer Learning[VGG]& N &  57.4\% & 0.39 \\
& Transfer Learning[VGG]& Y & 89.3\%\% & 0.83 \\
& Matching Networks& N & 80.5\% &0.3 \\
& Matching Networks& Y & 85.5\% &0.80 \\
& Matching Networks[miniImageNet]& N &84.7\%&0.63 \\
& Matching Networks[miniImageNet]& Y &91.4\%&0.80 \\
& SSM-Net& N&85.4\%&0.72\\
& \textbf{SSM-Net }& Y&\textbf{94.3\%}&\textbf{0.90} \\
\hline
\end{tabular}
\end{center}
\caption{Accuracy and  F1-Score performance of SSM Net, Matching Networks and Transfer Learning variants on Sugarcane disease data-set. Each
result is obtained over ~250 epochs. \textbf{Note:} All Matching Networks and SSM-Net Experiments outcomes are from 5-Way 5-Shot setup}
\end{table*}

 \begin{figure*}[!ht]
\begin{center}
% \fbox{\rule{0pt}{2in} \rule{0.4\linewidth}{0pt}
   \includegraphics[width=0.85\linewidth]{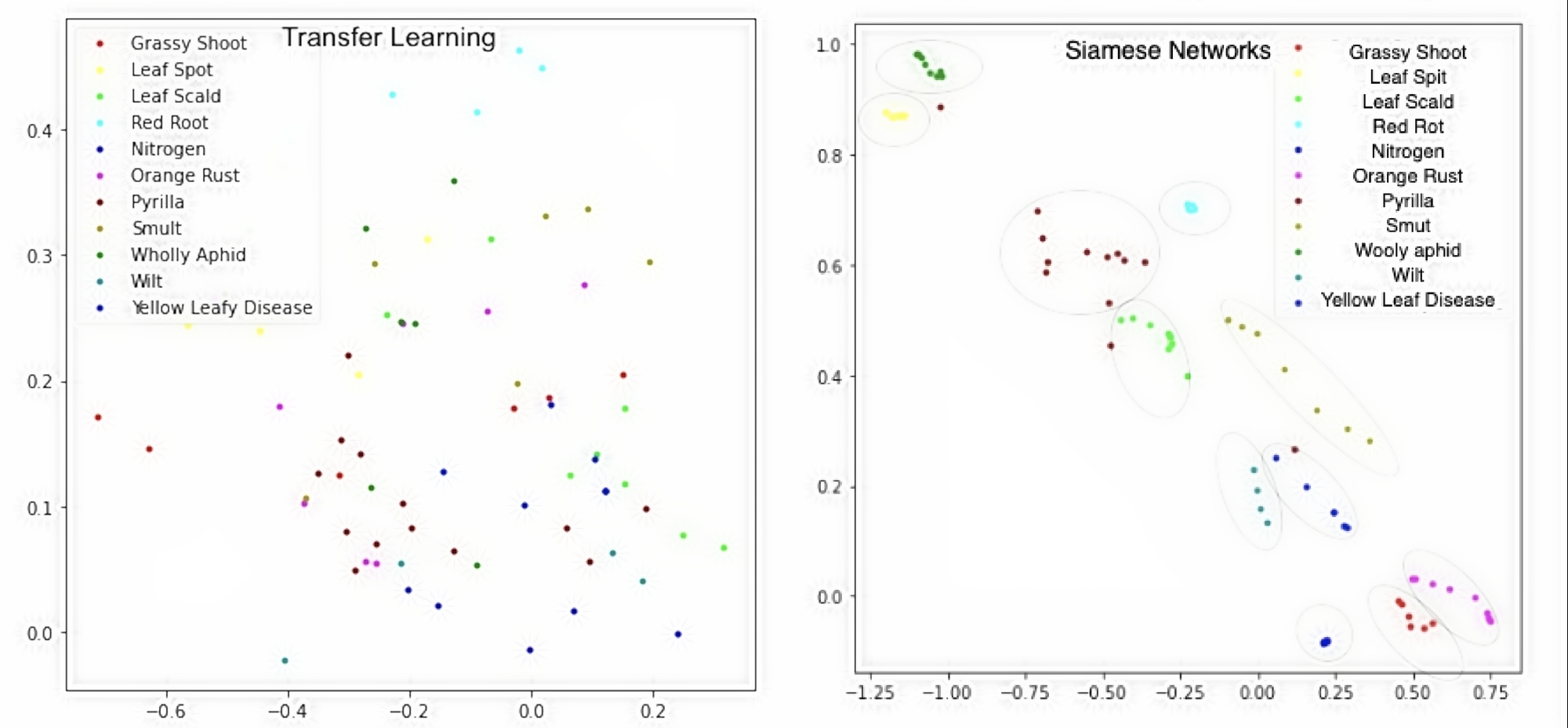}
\end{center}
   \caption{Feature Embeddings visualization of Transfer Learning vs Modified Siamese Network Embeddings on Sugarcane datset.}
\label{fig:6}
\label{fig:onecol}
\end{figure*}

% \begin{figure}
% \begin{center}
% % \fbox{\rule{0pt}{2in} \rule{0.4\linewidth}{0pt}
%   \includegraphics[width=1.0\linewidth]{image.png}
% \end{center}
%   \caption{Loss and Accuracy curve of SSM Networks. We can observe that as the number of epochs increases Validation Accuracy also increases.}
% \label{fig:8}
% \end{figure}

%------------------------------------------------------------------------
\section{Conclusion and Future Work}
Crop disease detection plays a crucial role in improving agricultural practices. If we can successfully automate early detection of crop disease, it will help us save on the amount of  pesticides used and reduce crop damage. Here, we have proposed a custom metrics-based few-shot learning method, SSM Net. In this, we leveraged transfer learning and metrics-based few-shot learning approaches to tackle the problem of low data disease identification. We showcased that: 
\begin{enumerate}
    \item With the help of combined transfer learning and Siamese networks, we can obtain better feature embeddings.
    \item Using SSM-Net we can achieve better accuracy in plants disease identification even with less amount of data.
\end{enumerate}
We envision that the proposed workflow might be applicable to other datasets \cite{Jadon_2020} that we will explore in the future. Our code implementation is available on Github: https://github.com/shruti-jadon/PlantsDiseaseDetection.

{\small
\bibliographystyle{ieee_fullname}
\bibliography{egbib}

\begin{thebibliography}{10}\itemsep=-1pt

\bibitem{aicrowd}
Minileaves: Challenges.

\bibitem{badage2018crop}
Anuradha Badage.
\newblock Crop disease detection using machine learning: Indian agriculture.
\newblock {\em IRJETV}, 2018.

\bibitem{badnakhe2011application}
Mrunalini~R Badnakhe and Prashant~R Deshmukh.
\newblock An application of k-means clustering and artificial intelligence in
  pattern recognition for crop diseases.
\newblock In {\em International Conference on Advancements in Information
  Technology}, volume~20, pages 134--138, 2011.

\bibitem{Bhatia2020}
Gresha~S. Bhatia, Pankaj Ahuja, Devendra Chaudhari, Sanket Paratkar, and
  Akshaya Patil.
\newblock Plant disease detection using deep learning.
\newblock In {\em Second International Conference on Computer Networks and
  Communication Technologies}, pages 408--415. Springer International
  Publishing, 2020.

\bibitem{cap2020leafgan}
Quan~Huu Cap, Hiroyuki Uga, Satoshi Kagiwada, and Hitoshi Iyatomi.
\newblock Leafgan: An effective data augmentation method for practical plant
  disease diagnosis.
\newblock {\em arXiv preprint arXiv:2002.10100}, 2020.

\bibitem{chaudhary2012color}
Piyush Chaudhary, Anand~K Chaudhari, AN Cheeran, and Sharda Godara.
\newblock Color transform based approach for disease spot detection on plant
  leaf.
\newblock {\em International journal of computer science and
  telecommunications}, 3(6):65--70, 2012.

\bibitem{hoppin2017pesticides}
Jane~A Hoppin, David~M Umbach, Stuart Long, Stephanie~J London, Paul~K
  Henneberger, Aaron Blair, Michael Alavanja, Laura E~Beane Freeman, and Dale~P
  Sandler.
\newblock Pesticides are associated with allergic and non-allergic wheeze among
  male farmers.
\newblock {\em Environmental health perspectives}, 125(4):535--543, 2017.

\bibitem{jadon2019hands}
S JADON.
\newblock Hands-on one-shot learning with python: A practical guide to
  implementing fast and... accurate deep learning models with fewer training,
  2019.

\bibitem{Jadon_2020}
Shruti Jadon, Owen~P. Leary, Ian Pan, Tyler~J. Harder, David~W. Wright, Lisa~H.
  Merck, and Derek Merck.
\newblock A comparative study of 2d image segmentation algorithms for traumatic
  brain lesions using {CT} data from the {ProTECTIII} multicenter clinical
  trial.
\newblock In Thomas~M. Deserno and Po-Hao Chen, editors, {\em Medical Imaging
  2020: Imaging Informatics for Healthcare, Research, and Applications}.
  {SPIE}, mar 2020.

\bibitem{Jadon2019ImprovingSN}
Shruti Jadon and Aditya Srinivasan.
\newblock Improving siamese networks for one shot learning using kernel based
  activation functions.
\newblock {\em ArXiv}, abs/1910.09798, 2019.

\bibitem{koch2015siamese}
Gregory Koch, Richard Zemel, and Ruslan Salakhutdinov.
\newblock Siamese neural networks for one-shot image recognition.
\newblock In {\em ICML deep learning workshop}, volume~2. Lille, 2015.

\bibitem{maniyath2018plant}
Shima~Ramesh Maniyath, PV Vinod, M Niveditha, R Pooja, N Shashank, Ramachandra
  Hebbar, et~al.
\newblock Plant disease detection using machine learning.
\newblock In {\em 2018 International Conference on Design Innovations for 3Cs
  Compute Communicate Control (ICDI3C)}, pages 41--45. IEEE, 2018.

\bibitem{Nagasubramanian2019}
Koushik Nagasubramanian, Sarah Jones, Asheesh~K. Singh, Soumik Sarkar, Arti
  Singh, and Baskar Ganapathysubramanian.
\newblock Plant disease identification using explainable 3d deep learning on
  hyperspectral images.
\newblock {\em Plant Methods}, 15(1), Aug. 2019.

\bibitem{russakovsky2015imagenet}
Olga Russakovsky, Jia Deng, Hao Su, Jonathan Krause, Sanjeev Satheesh, Sean Ma,
  Zhiheng Huang, Andrej Karpathy, Aditya Khosla, Michael Bernstein, et~al.
\newblock Imagenet large scale visual recognition challenge.
\newblock {\em International journal of computer vision}, 115(3):211--252,
  2015.

\bibitem{Saleem2019}
Saleem, Potgieter, and Mahmood Arif.
\newblock Plant disease detection and classification by deep learning.
\newblock {\em Plants}, 8(11):468, Oct. 2019.

\bibitem{simonyan2014very}
Karen Simonyan and Andrew Zisserman.
\newblock Very deep convolutional networks for large-scale image recognition.
\newblock {\em arXiv preprint arXiv:1409.1556}, 2014.

\bibitem{singh2018pesticide}
Ngangbam~Sarat Singh, Ranju Sharma, Talat Parween, and PK Patanjali.
\newblock Pesticide contamination and human health risk factor.
\newblock In {\em Modern Age Environmental Problems and their Remediation},
  pages 49--68. Springer, 2018.

\bibitem{vinyals2016matching}
Oriol Vinyals, Charles Blundell, Timothy Lillicrap, Daan Wierstra, et~al.
\newblock Matching networks for one shot learning.
\newblock In {\em Advances in neural information processing systems}, pages
  3630--3638, 2016.

\bibitem{Wang2017}
Guan Wang, Yu Sun, and Jianxin Wang.
\newblock Automatic image-based plant disease severity estimation using deep
  learning.
\newblock {\em Computational Intelligence and Neuroscience}, 2017:1--8, 2017.

\bibitem{weiss2016survey}
Karl Weiss, Taghi~M Khoshgoftaar, and DingDing Wang.
\newblock A survey of transfer learning.
\newblock {\em Journal of Big data}, 3(1):9, 2016.

\bibitem{wong2016understanding}
Sebastien~C Wong, Adam Gatt, Victor Stamatescu, and Mark~D McDonnell.
\newblock Understanding data augmentation for classification: when to warp?
\newblock In {\em 2016 international conference on digital image computing:
  techniques and applications (DICTA)}, pages 1--6. IEEE, 2016.

\end{thebibliography}
}

\end{document}